\pdfoutput=1
\documentclass[11pt]{article}

\usepackage{ACL2023}
\usepackage{booktabs}
\usepackage{xcolor}
\usepackage{colortbl}
\usepackage{microtype}
\newcommand{\ra}[1]{\renewcommand{\arraystretch}{#1}}
\usepackage{times}
\usepackage{latexsym}
\usepackage{caption}
\usepackage{tabularx}

\usepackage[T1]{fontenc}
\usepackage[utf8]{inputenc}

\usepackage{microtype}

\usepackage{inconsolata}

\title{What Makes a Good Paraphrase: Do Automated Evaluations Work? }

\author{Anna Moskvina, Bhushan Kotnis, Chris Catacata, Michael Janz \and Nasrin Saef\\
  Ella Lab – Gesellschaft für künstliche Intelligenz mbH\\
  Cologne \\
  Germany \\
  \texttt{\{anna.moskvina,bhushan.kotnis,chris.catacata,michael.janz,nasrin.saef\}@ella-lab.io} %\\\And
  \\}

\begin{document}
%This document is a supplement to the general instructions for *ACL authors. It contains instructions for using the \LaTeX{} style file for ACL 2023.
%The document itself conforms to its own specifications, and is, therefore, an example of what your manuscript should look like.
%These instructions should be used both for papers submitted for review and for final versions of accepted papers.

%max. 2 pages plus references

\maketitle
\begin{abstract}
Paraphrasing is the task of expressing an essential idea or meaning in different words. But how different should the words be in order to be considered an acceptable paraphrase? And can we exclusively use automated metrics to evaluate the quality of a paraphrase? We attempt to answer these questions by conducting experiments on a German data set and performing automatic and expert linguistic evaluation.
%Paraphrasing is the task of expressing an essential idea or meaning in different words. In order to be considered an acceptable paraphrase how different should the words be? And can we use only automated metrics to evaluate paraphrase quality? We attempt to answer these questions by conducting experiments on German data set and performing automatic and expert linguistic evaluation.
%This paper is dedicated to the issue of identifying "good" paraphrases extracted from a german data set. We tackle the problem of what should be considered as an acceptable paraphrase pair from several angles: quantitatively with automated scoring metrics and qualitatively (from a linguistic point of view).
\end{abstract}

\section{Introduction}
One of the most beautiful features of human language is that it allows one to express the same idea in diverse ways. A paraphrase is exactly that, the same idea expressed in different words. For example, the sentence "\textit{How can I learn to write better paraphrases?}" can be paraphrased as "\textit{How do I improve my paraphrasing skills?}" Paraphrases are essentially meaning preserving alterations of the original text. These alterations or transformations may be based on speech register, diction level, style and compression \cite{Clarke2008,Yatskar2010,Xu2012}. The paraphrasing task has received a lot of attention in the form of many paraphrase data sets such as PPDB \cite{Ganitkevitch2013}, Wikianswer \cite{fader2013} and MSCOCO \cite{lin2014} among others. This has led to the development of various rule based and neural models \cite{zhou2021}. But only few studies exist that aim to answer the question: what makes a good paraphrase from the perspective of a human reader? Additionally, the data sets mentioned above are in English.
 %However the notion of a good paraphrase from the perspective of the reader may differ across languages and this is not well studied. 
Here we aim to shed some light on this problem based on our experience in creating paraphrases for the German language. 

More specifically we study two different paradigms of evaluating paraphrases: automated and linguistic. We then test these metrics by conducting experiments on the existing large German corpus GC4\footnote{https://german-nlp-group.github.io/projects/gc4-corpus.html}.

\section{Methodology}
To estimate how a paraphrase pair can be assessed, we suggest using a combination of several automated metrics with the linguistic analysis. To create a new German paraphrase data set that is based on genuine news articles and blogs, we extracted paraphrase candidates from a subset of 50.000 articles from GC4 data set as suggested by \citet{dong2021parasci}.
%\textbf{Paraphrase extraction:} We used the GC4 Dataset as it is the largest currently available collection of news articles, blogs, job descriptions etc. 

As a part of our preprocessing framework we filtered sources based on country, keeping only those articles and blog entries that originated in Germany. Though newspapers from Austria, Switzerland and other German-speaking countries can also be a rich source of data, for this experiment we omit using regional dialects. Additionally, any sources that had less than 1000 articles were removed: we hypothesize that sources with little material would not have many different versions depicting the same events and would be a poor choice for paraphrase extraction. Furthermore, we remove articles that had less than 700 characters, or did not end with a final punctuation ("!?."). Finally, we tokenized, normalized and cleaned the text by filtering out non relevant editorial information (such as bylines, contact information, calls to action, advertisements etc.).

Subsequently, we split the resulting texts into sentences and compare each sentence to all of the other sentences using cosine similarity of Bert Sentence embeddings \cite{reimers2019sentence}. This is done to measure the semantic closeness of the paraphrase candidates. If the cosine similarity is higher than 0.935, we consider this pair as a potential paraphrase pair. This results in 96.066 paraphrase candidates. These candidates are then subjected to automated paraphrase evaluation scores in the first round of evaluation.

\section{Paraphrase Evaluation}
The paraphrase candidates are evaluated in two rounds: with automated scores and by an expert linguist. After the first round we filter the candidates based on the acquired scores. Only this filtered subset of original paraphrase candidates is evaluated manually.

\begin{table*}[t]
\centering
\ra{1.1}
{\tiny
\begin{tabular}{@{}p{1em}p{4.0cm}p{4.0cm}p{1em}llllllll@{}}
\toprule
\textbf{Num.} &
\textbf{Original Sentence} &
  \textbf{Paraphrased Sentence} &
  \textbf{FCS} &
  \textbf{GS} &
  $\mathbf{R_1}$ &
  $\mathbf{R_2}$ &
  $\mathbf{R_n}$ &
  \textbf{BS} & \textbf{LD}
   & \textbf{CS} 
   & \textbf{OQ}
   \\ \midrule
   1 &
Die Polizei sprach von einem Schaden in Millionenhöhe &
  Die Polizei spricht von einem Millionenschaden &
  1.0 &
  1.0 &
  0.63 &
  0.35 &
  0.0 & 0.88
   & 0.40
   & 0.70 & 0.50
   \\
   2 &
Die Zahl der Jungen und Mädchen, die keine Grundschule besuchen, stagniert. &
  Und die Zahl der Kinder, die keine Grundschule besuchen, stagniert. &
  1.0 &
  1.0 &
  0.46 &
  0.18 &
  0.0 & 0.93
   & 0.30
   & 0.90 & 0.75
   \\
   3 &
Der Ausstoß klimaschädlicher Treibhausgase soll bis 2030 um mindestens 40 Prozent verglichen mit 1990 sinken. &
  Geplant sind 40 Prozent weniger Treibhausgasemissionen bis 2020 im Vergleich zu 1990. &
  0 &
  1.0 &
  0.74 &
  0.54 &
  0.33 & 0.79
   & 0.75
   & 0.60 & 0.60
   \\
   4 &
Volkswagen erklärte bisher, sich an alle gültigen Regeln gehalten zu haben. &
  Volkswagen ist der Überzeugung, alle Regeln eingehalten zu haben. &
  1.0 &
  1.0 &
  0.57 &
  0.17 &
  0.0 & 0.83
   & 0.80
   & 0.95 & 0.95
   \\
   5 &
Die sofort verständigte Feuerwehr brachte den Brand schnell unter Kontrolle. &
  Den Brand hatten die Feuerwehrleute schnell unter Kontrolle. &
  1.0 &
  1.0 &
  0.53 &
  0.37 &
  0.19 & 0.85
   & 0.70
   & 0.95 & 0.85
   \\
   6 &
Wann mit den Baumaßnahmen konkret begonnen werden kann, ist allerdings noch offen. &
  Noch steht aber nicht fest, wann es mit dem Bauen losgeht. &
  1.0 &
  1.0 &
  0.77 &
  0.61 &
  0.50 & 0.86
   & 0.70
   & 0.95 & 0.85
   \\
   7 &
Der Kunstbegriff wurde aus den englischen Worten für Griechenland (Greece) und Ausstieg (Exit) gebildet – gemeint ist ein Ausstieg oder Rauswurf Griechenlands aus der Eurozone. &
  Das Wort setzt sich aus Greece und exit zusammen und meint das Ausscheiden Griechenlands aus der Eurozone. &
  0.0 &
  0.65 &
  0.80 &
  0.62 &
  0.40 & 0.79
   & 0.60
   & 0.80 & 0.70
   \\
   8 &
Das Feuer war am späten Donnerstagabend in einer Wohnung eines Mehrfamilienhauses ausgebrochen. &
  Das Feuer war am späten Montagabend in einer Wohnung ausgebrochen. &
  1.0 &
  1.0 &
  0.50 &
  0.11 &
  0.0 & 0.90
   & 0.50
   & 0.60 & 0.50
   \\
   9 &
Aber im Grunde ist das auch egal. &
  Aber das spielt keine Rolle. &
  1.0 &
  1.0 &
  0.78 &
  0.57 &
  0.42 & 0.81
   & 0.85
   & 0.90 & 0.90
   \\ 
   10 &
Bei der Tat entstand ein Schaden in Höhe von mehreren hundert Euro. &
  Der Schaden beliefe sich sogar auf eine Million Euro. &
  1.0 &
  1.0 &
  0.58 &
  0.27 &
  0.15 & 0.83
   & 0.40
   & 0.10 & 0.10
   \\
  \bottomrule

\end{tabular}
\caption{Examples of paraphrases evaluated with automated scores : factual correctness (FCS), grammar (GS), Rouge Unigram ($R_1$), Rouge Bigram ($R_2$), Rouge Longest Ngram ($R_n$), BERTScore (BS) and human evaluations: Lexical Diversity (LD), Content Similarity (CS) and Overall Quality (OQ)
}
\label{tab:examples}}
\end{table*}

\subsection{Automated Paraphrase Evaluation}

%\bhushan{Bhushan: In my view we don't have enough space to describe these in detail. In my view we should first state the results and then weave explanations inside the discussion.}
%\anja{I was thinking that the Discussion part can be the subsection called Linguistic paraphrase evaluation, otherwise we do not have spacy at all for it. Remember, that we have only 2 pages. And when I talked to @Nasrin, she mentioned that we should explain the scores themselves a little bit.}
As a part of the automated evaluation we calculate standard scores such as Rouge Score, Bert Score and Grammar Score as well as the Fact Check Metric introduced by us in this articles:
\begin{itemize}
    \item \textbf{Rouge Scores} for unigram, bigram and longest n-gram \cite{lin2004rouge}. Though these scores were originally designed for evaluating automatic summarization, they can also be adopted for comparing paraphrase pairs. $R_1$ computes the F1-score of the matching unigrams, $R_2$ does the same for bigrams and $R_n$ for the longest n-gram. Higher scores indicate a greater syntactical overlap between the candidate and the paraphrase.
    \item \textbf{BertScore} \cite{zhang2019bertscore} to calculate the closeness of a candidate pair, based on cosine distance between contextual embeddings obtained from BERT.%, followed by F-1 score.
    \item \textbf{Factual Correctness Score} (FCS) to measure the number of "facts" present in both sentences. "Facts" are any numbers, dates and entities found with NER provided by spaCy\footnote{ https://spacy.io/api/entityrecognizer}. We use exact matches to compute the factual correctness score. If in an original sentence a person’s name was written as "John Doe" and in the paraphrase candidate as "Mr. Doe", then the entity would be regarded as "missing" in the second sentence.\footnote{ Those limitations are to be addressed in future work.}.
    \item \textbf{Grammar Score} (GS) to check errors in spelling and grammar facilitated by Language Tool\footnote{ https://languagetool.org/spellchecking-german}. The lower the score, the more grammatically correct the sentences are.
\end{itemize}

All of the scores have a complementary nature to one another. Using all of them in combination allows for a better understanding of the quality of a paraphrase pair and is helpful to establish a threshold for filtering candidates that are either too close to each other (variation in spelling or in the number of entities) or too different from each other (contain lexical changes that alter the meaning of the sentence completely, for instance, with negation).

\subsection{Linguistic Paraphrase Evaluation}
The second evaluation is conducted by linguistic experts. 7 reviewers rate different aspects (listed below) of a paraphrase pair on a scale from 1 (worst) to 5 (best). Each aspect is formulated as a statement and reviewers have to answer whether they agree with it or not. As a result this metric is subjective. To counter this subjectivity the ratings are normalized.

\begin{itemize}
    \item \textbf{Linguistic Difference} (LD): a measurement of how different the surface form of the paraphrase is compared to its source text.
    \item \textbf{Content Similarity} (CS): a measurement of how close the semantic and pragmatic content of the paraphrase is to its source text.
    \item \textbf{Overall Quality} (OQ): a measurement based on the following definition of a paraphrase: “restating of a sentence such that both sentences would generally be recognized as lexically and syntactically different while remaining semantically equal” \cite{mccarthy2009components}.

\end{itemize}

Though the reviewers are trained to evaluate paraphrases and we always use aggregated, normalized results, the results of these metrics are subjective to some extent.

\section{Discussion}
 % straightforward correlation between automatic and linguistic scores. 
In Table~\ref{tab:examples} we provide a few samples for evaluation and discussion. In these examples we do not see a correlation between automated and linguistic scores. Indeed, based on the automated scores such as \textbf{FCS}, $\mathbf{R_2}$, \textbf{GS} one would assume that example \textbf{10}s an acceptable paraphrase pair. This is not the case, however, because the reviewers gave it a very low \textbf{OQ} score (0.10). This inconsistency in the scores can be attributed to fact-checking, which failed to identify the change from "Millionen" ("millions") to "hundert" ("hundred"). This suggests that automated scores are not always reliable.

However, it is also possible that higher values of automated metrics may coincide with high reviewer scores. This is seen in example \textbf{4} with low \textbf{Rouge scores} (lower overlap) and high \textbf{FCS} and \textbf{GC} scores corresponding with high reviewer scores (\textbf{OQ}>0.90). It is important to note that the linguistic evaluation always takes place after the potential candidates are filtered out.

These findings emphasize the complementary nature of the evaluations and the importance of combining both types of evaluation.

\section{Future Work}
%\bhushan{Bhushan: I doubt we will have space for future work, given that we need to add results and the discussion/conclusion.}
%\anja{no problem}
As part of the ongoing research we plan on investigating the following questions: why are some of the texts more difficult to paraphrase than others by a large language model (LLM)? Is it governed by a linguistic factor or an external factor (data the LLM has been trained on)? Additionally, we intend to enrich the current evaluation scores and tackle the issue of fuzzy factual correctness.

\section{Acknowledgment}
This study was conducted within the framework of developing a commercial paraphrasing tool for German by employees of Ella Lab – Gesellschaft für künstliche Intelligenz mbH.

\bibliography{anthology,custom}

\begin{thebibliography}{12}
\expandafter\ifx\csname natexlab\endcsname\relax\def\natexlab#1{#1}\fi

\bibitem[{Clarke and Lapata(2008)}]{Clarke2008}
James Clarke and Mirella Lapata. 2008.
\newblock Global inference for sentence compression: An integer linear
  programming approach.
\newblock \emph{Journal of Artificial Intelligence Research}, 31:399--429.

\bibitem[{Dong et~al.(2021)Dong, Wan, and Cao}]{dong2021parasci}
Qingxiu Dong, Xiaojun Wan, and Yue Cao. 2021.
\newblock Parasci: A large scientific paraphrase dataset for longer paraphrase
  generation.
\newblock \emph{arXiv preprint arXiv:2101.08382}.

\bibitem[{Fader et~al.(2013)Fader, Zettlemoyer, and Etzioni}]{fader2013}
Anthony Fader, Luke Zettlemoyer, and Oren Etzioni. 2013.
\newblock Paraphrase-driven learning for open question answering.
\newblock In \emph{Proceedings of the 51st Annual Meeting of the Association
  for Computational Linguistics (Volume 1: Long Papers)}, pages 1608--1618.

\bibitem[{Ganitkevitch et~al.(2013)Ganitkevitch, Van~Durme, and
  Callison-Burch}]{Ganitkevitch2013}
Juri Ganitkevitch, Benjamin Van~Durme, and Chris Callison-Burch. 2013.
\newblock Ppdb: The paraphrase database.
\newblock In \emph{Proceedings of the 2013 Conference of the North American
  Chapter of the Association for Computational Linguistics: Human Language
  Technologies}, pages 758--764.

\bibitem[{Lin(2004)}]{lin2004rouge}
Chin-Yew Lin. 2004.
\newblock Rouge: A package for automatic evaluation of summaries.
\newblock In \emph{Text summarization branches out}, pages 74--81.

\bibitem[{Lin et~al.(2014)Lin, Maire, Belongie, Hays, Perona, Ramanan,
  Doll{\'a}r, and Zitnick}]{lin2014}
Tsung-Yi Lin, Michael Maire, Serge Belongie, James Hays, Pietro Perona, Deva
  Ramanan, Piotr Doll{\'a}r, and C~Lawrence Zitnick. 2014.
\newblock Microsoft coco: Common objects in context.
\newblock In \emph{Computer Vision--ECCV 2014: 13th European Conference,
  Zurich, Switzerland, September 6-12, 2014, Proceedings, Part V 13}, pages
  740--755. Springer.

\bibitem[{McCarthy et~al.(2009)McCarthy, Guess, and
  McNamara}]{mccarthy2009components}
Philip~M McCarthy, Rebekah~H Guess, and Danielle~S McNamara. 2009.
\newblock The components of paraphrase evaluations.
\newblock \emph{Behavior Research Methods}, 41(3):682--690.

\bibitem[{Reimers and Gurevych(2019)}]{reimers2019sentence}
Nils Reimers and Iryna Gurevych. 2019.
\newblock Sentence-bert: Sentence embeddings using siamese bert-networks.
\newblock \emph{arXiv preprint arXiv:1908.10084}.

\bibitem[{Xu et~al.(2012)Xu, Ritter, Dolan, Grishman, and Cherry}]{Xu2012}
Wei Xu, Alan Ritter, William~B Dolan, Ralph Grishman, and Colin Cherry. 2012.
\newblock Paraphrasing for style.
\newblock In \emph{Proceedings of COLING 2012}, pages 2899--2914.

\bibitem[{Yatskar et~al.(2010)Yatskar, Pang, Danescu-Niculescu-Mizil, and
  Lee}]{Yatskar2010}
Mark Yatskar, Bo~Pang, Cristian Danescu-Niculescu-Mizil, and Lillian Lee. 2010.
\newblock For the sake of simplicity: Unsupervised extraction of lexical
  simplifications from wikipedia.
\newblock \emph{arXiv preprint arXiv:1008.1986}.

\bibitem[{Zhang et~al.(2019)Zhang, Kishore, Wu, Weinberger, and
  Artzi}]{zhang2019bertscore}
Tianyi Zhang, Varsha Kishore, Felix Wu, Kilian~Q Weinberger, and Yoav Artzi.
  2019.
\newblock Bertscore: Evaluating text generation with bert.
\newblock \emph{arXiv preprint arXiv:1904.09675}.

\bibitem[{Zhou and Bhat(2021)}]{zhou2021}
Jianing Zhou and Suma Bhat. 2021.
\newblock Paraphrase generation: A survey of the state of the art.
\newblock In \emph{Proceedings of the 2021 Conference on Empirical Methods in
  Natural Language Processing}, pages 5075--5086.

\end{thebibliography}
\bibliographystyle{acl_natbib}
\end{document}